\documentclass[sigconf]{acmart}

\copyrightyear{2022}
\acmYear{2022}
\setcopyright{acmcopyright}
\acmConference[ICTIR '22] {Proceedings of the 2022 ACM SIGIR International Conference on the Theory of Information Retrieval}{July 11--12, 2022}{Madrid, Spain.}
\acmBooktitle{Proceedings of the 2022 ACM SIGIR International Conference on the Theory of Information Retrieval (ICTIR '22), July 11--12, 2022, Madrid, Spain}
\acmPrice{15.00}
\acmISBN{978-1-4503-9412-3/22/07}
\acmDOI{10.1145/3539813.3545144}

\usepackage{tabularx}
\usepackage{bm}
\usepackage[textsize=footnotesize]{todonotes} 
\usepackage{multirow}
\usepackage{arydshln}

\usepackage{pifont}
\usepackage{diagbox}
\usepackage{tcolorbox}
\usepackage{lipsum}  

\begin{document}

\title{The Role of Complex NLP in Transformers for Text Ranking?}

\author{David Rau}
\affiliation{
  \institution{University of Amsterdam}
  \city{Amsterdam}
  \country{The Netherlands}
}
\author{Jaap Kamps}
\affiliation{
  \institution{University of Amsterdam}
  \city{Amsterdam}
  \country{The Netherlands}
}

\renewcommand{\shortauthors}{D. Rau \and J. Kamps}

\begin{abstract}
Even though term-based methods such as BM25 provide strong baselines in ranking, under certain conditions they are dominated by large pre-trained masked language models (MLMs) such as BERT.  To date, the source of their effectiveness remains unclear.
Is it their ability to truly understand the meaning through modeling syntactic aspects?
We answer this by manipulating the input order and position information in a way that destroys the natural sequence order of query and passage and shows that the model still achieves comparable performance. Overall, our results highlight that syntactic aspects do not play a critical role in the effectiveness of re-ranking with BERT. We point to other mechanisms such as query-passage cross-attention and richer embeddings that capture word meanings based on aggregated context regardless of the word order for being the main attributions for its superior performance.
 
\end{abstract}



\maketitle

\section{Introduction}
Originating from the field of natural language processing (NLP) large-scale self-supervised training yields representations that are useful for a wide range of tasks \cite{delvinbert, wang2019glue, Nogueira2019PassageRW}. Specifically, pre-training large language models using masked language modeling (MLM) as proposed by \cite{delvinbert} in BERT has become a standard procedure to achieve top performances on downstream tasks. 

While in the past many ideas coming from NLP did not lead to convincing improvements in information retrieval \cite{fagan1988experiments, arampatzis1998phase, wsd1, wsd2}, somewhat surprisingly BERT did lead to the long-awaited jump in performance (see also \cite{linneuralhype}).
Nevertheless, its success comes with the caveat of extremely complex models that are hard to interpret, and therefore it is hard to pinpoint the source of their effectiveness. These rankers typically comprise millions of parameters requiring massive amounts of training data. Only with the arrival of the large-scale ranking dataset MS MARCO~\citep{bajaj2016ms}, did large MLMs find their successful application in information retrieval.

\begin{table}
    \centering
\begin{tcolorbox}[colback=white,boxrule=0.5pt,left=1pt,right=1pt,top=1pt,bottom=1pt]
\small
\begin{tabularx}{\linewidth}{lX}
\textit{Original Query} & what the best way to get clothes white \\
\textit{Original Passage} & bleach is also the best way to get white clothes white again, and helps remove stubborn, older stains. choose warm water for moderately soiled, synthetic blend white clothes wash white clothes in warm water if they're moderately soiled, are lined, and if they're made of synthetic fibers or natural and synthetic blends. \\
& \textbf{Predicted Rank: 1} \\
\midrule
\textit{Shuffled Query} & clothes white best get way what to the\\
\textit{Shuffled Passage} & stains \#\#lea moderately moderately fibers stubborn blend blend synthetic synthetic synthetic wash lined helps remove soil soil choose clothes clothes clothes warm warm older natural \#\#ch white white white white water water best again get re re way \#\#ed \#\#ed made if if also or they they are \#\#s for is to in and and and of the b . . , , , , , ' ' \\
& \textbf{Predicted Rank: 1} \\
\end{tabularx}
\end{tcolorbox}
    \caption{The same query (id: 1108651) and passage (id: 8175412) in original and perturbed order. While breaking the natural sequence order makes it meaningless to human readers, a model trained on the perturbed sequence estimates its relevance correct. Example taken from the NIST 2020 testset on MSMARCO.}
        \label{tab:example}
\end{table}

\begin{figure}
	\includegraphics[width=0.7\linewidth]{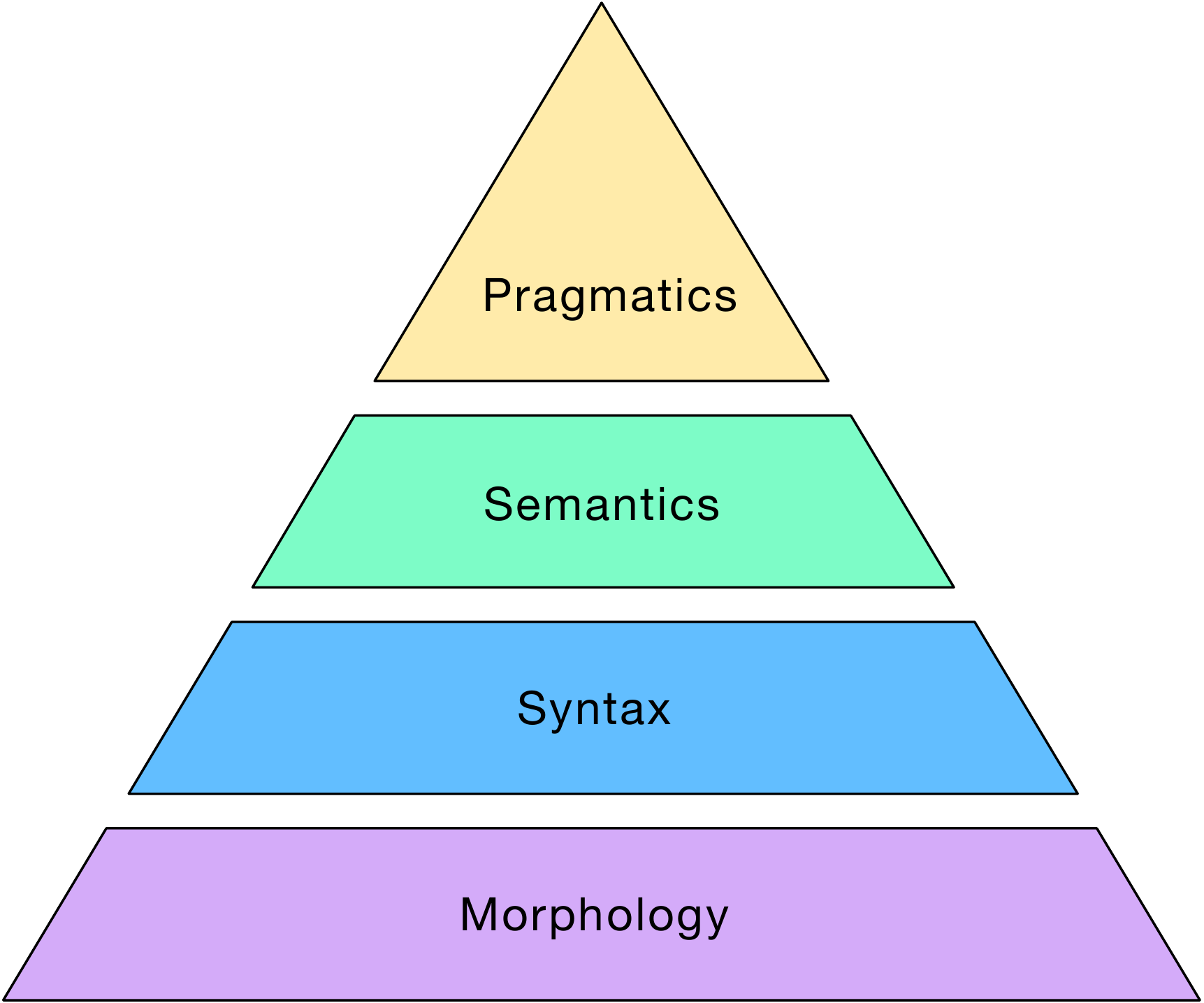}
	\caption{NLP pyramid. Four hierarchical steps to describe the understanding of natural language.}
	\label{fig:nlp_pyramid}
\end{figure}

With BERT's representation of long sequences of input tokens, providing the means to model syntactic aspects of the input, information retrieval seems to shift towards being an NLP problem. 	
To recap basic linguistics, understanding natural language can be comprehended in four hierarchical steps. The NLP pyramid (Figure~\ref{fig:nlp_pyramid}) depicts the consecutive steps from bottom up. It starts on the word level with the morphology describing how words are formed depending on their context (singular/plural, word inflection, etc. ). The next step considers the relationship between multiple words building the syntax of a sentence.  
Through the structure of a sentence, we can %
understand the function of words (parts-of-speech), identify sentence boundaries, and understand the dependency between words. With having an understanding of the syntax we can then derive the semantics or ``meaning" of a sentence. As the last step, pragmatics describes the higher level of semiotics and spans the text as a whole. In case the sequence embeddings of BERT are capturing some of these higher levels, this could explain the observed gains in ranking effectiveness observed in ranking tasks.

We want to focus on the second step, the syntax, and observe that here the order of words in the sentence is of essential importance.  In most cases altering the order changes the meaning or even destroys the grammar making the meaning of the sentence undefined.  Applied to the ranking problem: consider the example query and document in Table~\ref{tab:example}. For humans it is easy to identify the passage in natural sequence order (top half) as relevant to the query, however, with perturbed sequence order (bottom half) the passage becomes meaningless and therefore it is very hard if not impossible to estimate its relevance. 
	
Remarkably, traditional rankers such as Query likelihood and BM25 pose such strong baselines by solely operating on bag-of-words representations that disregard the sequence order and therewith the syntax entirely. One of the key differences of BERT over these lexical rankers is the ability to go beyond the word level and to be able to model phrase and sentence level contexts. While such large models could potentially gain deeper semantic and syntactic abstractions to understand the true meaning of documents \cite{dehghanitoward,kratzer1998semantics} it remains unclear whether they do so. Little is known about how BERT estimates the relevance of a query-document pair; what features are encoded and which of those are essential for its performance. 

One possible explanation for the success of BERT is that it indeed learns to understand syntax. Applied to ranking, BERT could potentially build deep interactions between queries and documents that allow uncovering complex relevance patterns bringing us one step closer to the vision for future retrieval systems of \citet{metzler2021} in ``Making Domain Experts out of Dilettantes''. In contrast to this, another possible explanation could be that BERT lines up alongside other NLP techniques \cite{deerwester1990indexing,mikolov2013distributed, peters} exploiting the distributional properties of natural language \cite{harris} by merely learning simple term distributions.

In this paper, we overall aim to answer the question: 
\begin{description}
	\item \textit{How much is modeling syntactic aspects contributing to the success of BERT in information retrieval?}
\end{description}

 We organize our paper by answering the following research questions:

\newcommand{\rqa}{Does the BERT Cross-Encoder Ranker perform well due to modeling syntactic aspects?}

\newcommand{\rqb}{
Does the model inherently lose order sensitivity while fine-tuning on the ranking task? }

\newcommand{\rqc}{
How different are models that are trained without sequence order?}

\newcommand{\rqd}{What is the effect of removing position information on the ranking performance?}

\newcommand{\rqe}{What role does modeling syntactic aspects play in Natural Language Understanding tasks?}

\begin{description}
\item[\textbf{RQ1}] \sl \rqa
\end{description}

\begin{description}
\item[\textbf{RQ2}] \sl \rqd
\end{description}

\begin{description}
\item[\textbf{RQ3}] \sl \rqb
\end{description}

\begin{description}
\item[\textbf{RQ4}] \sl \rqc
\end{description}

\begin{description}
\item[\textbf{RQ5}] \sl \rqe
\end{description}

To this end, we are manipulating the natural order, or how the model perceives order in different ways. 
More specifically, in Section \ref{sec:perturbing} we conduct experiments manipulating the sequence order during \emph{fine-tune training} and fine-tuning evaluation. 
In Section \ref{sec:nopos} we alter the BERT Cross-Encoder in a way that it can not perceive sequence order anymore during fine-tuning (creating a true BOW-BERT). 
We later, in Section \ref{sec:analysis} analyze the models that have been rendered order invariant by comparing their latent representations to the vanilla Cross-Encoder Ranker and the pre-trained BERT model.
Finally, in Section \ref{sec:nlu} we evaluate our manipulations on the basic NLP tasks using GLUE.

We find that for re-ranking with the BERT Cross-Encoder the sequence order does not play a critical role. Our main contribution lies in showing that the superior performance of the BERT ranker cannot be attributed to syntactic abstraction and a deeper understanding of language. We validate our findings on two tasks (re-ranking and NLU) and several datasets (MSMARCO, Robust04, and GLUE).

\smallskip
For reproducibility and to encourage further research in this direction, we make our new order-invariant BOW-BERT model available to the public under \url{https://github.com/davidmrau/ictir22}.

\section{Related Work}
In this section, We discuss related work from IR and NLP on probing large pre-trained transformers.

\medskip
Despite several efforts to open up the BERT ranker as a black-box~\citep{Formal2021AWB,rau2022different, camara2020diagnosing} the mechanisms within the model remain unclear. More particular to ranking, the role of word-order in large MLMs during inference has been studied recently \citep{macavaney2020abnirml,rennings2019axiomatic}. In previous work,
\citet{whatbertisnot} showed that BERT relies on the word order during pre-training for the MLM task, finding evidence of the model encoding syntactic information in its representations. The necessity of doing so, however, seems to vary depending on the task. One of the earlier works in NLP showing that the input order does not play a crucial role in solving natural language inference (SNLI) was carried out by \citet{decomposableattention} based on LSTMs. 

In later work, \cite{SinhaPPW20, pham-etal-2021-order, gupta2021bert} investigate the sensitivity of input order permutations during evaluation and find a varying order of sensitivity for some NLP down-stream tasks. \citet{SinhaPPW20} show that different transformer models perform well on permuted input of natural language inference tasks. \citet{gupta2021bert} show the same for RoBERTa \cite{roberta} on several natural language understanding tasks (MNLI, QQP and SST-2). \citet{pham-etal-2021-order} conduct a more complete study suggesting that some tasks are more sensitive to the input order than others based on the GLUE benchmark.  
However, recent work signals that the role of word order is smaller than expected on NLP tasks \citet{sinha-etal-2021-masked}. They carry out a wide range of experiments on order permutations on GLUE using RoBERTa. \citet{wang2021on} and \citet{sinha-etal-2021-masked} study the effect of removing the position information for a subset of the GLUE tasks.

As the work by \citet{sinha-etal-2021-masked} was carried in parallel to ours, a sub-set of the experiments on GLUE show similarities to ours, with a few differences: We additionally choose to break the sentence structure deterministically by sorting the input by id and to carry out the experiments on all tasks of the GLUE benchmark. 
Furthermore, previous work has examined sequence order on a sentence level, whereas ranking operates on a passage/document level.  In terms of text understanding this is posing quite another challenge. 
Hence, compared to previous work, we are the first to examine the role of syntactic aspects in ranking with large MLMs during learning the ranking task. 
Note that the analysis of the role of syntactic structures in information retrieval is of particular importance, as one of its main achievements was the development of effective ``bag of words'' approaches over 50 years ago, and these models are still powering almost every single search index deployed in practice.

\section{Experimental Setup}
In this section, we detail the basic neural ranker and two test collections used in the experiments.

\subsection{Model}
For our experiments we use a  BERT Cross-Encoder (CE) which encodes both queries and passages at the same time. Given input:
\begin{equation*}
	 \bm{x} \in \{[CLS], q_1, \dots, q_n\, [SEP], p_1, \dots , p_m, [SEP]\},
\end{equation*}
where $q$ represents query tokens and $p$ passage tokens, the activations of the CLS token are fed to a binary classifier layer to classify a passage as relevant or non-relevant; the relevance probability is then used as a relevance score to re-rank the passages.

\subsection{Data}
\subsubsection{MSMARCO}
We conduct our ranking experiments on the TREC 2020 Deep Learning Track's passage retrieval task on the MS MARCO dataset~\cite{bajaj2016ms}. The average passage length is 56. For training, we use the official MS MARCO training triplets. For evaluation, we use the NIST 2020 judgments with the official top-1,000 runs. We evaluate using the metrics NDCG@10, MAP, and Recall@100.

\subsubsection{Robust04}
TREC 2004 Robust Track is a news collection of documents with an average length of 254 words. We use this collection as a \emph{zero shot} re-ranking test collection with no additional training on the dataset. We choose this additional dataset for document retrieval due to its complete judgments allowing for unbiased evaluation. For evaluation we use the title queries 301-450 and 601-700. BM25 with no stemming serves as a first stage ranker to retrieve the top-1,000 ranks for each topic. Evaluation is done using the metrics NDCG@10, MAP, and Recall@100.

\subsection{Training Details}
We follow the training scheme of \citet{Nogueira2019PassageRW} unless stated differently. 
We train our BERT ranker from Huggingface's~\cite{ wolf-etal-2020-transformers} \textit{bert-base-uncased}. For training we use batch size 64, maximum sequence length 512,  warm-up steps 1000, learning rate 3e-6, epoch size 1000 and evaluate the best model within 40 epochs.

\section{Perturbing Sequence Order}
\label{sec:perturbing}
In this section, we conduct experiments to examine the role of syntax understanding in the performance gain of the BERT Cross-Encoder for passage re-ranking. 

\medskip
Ranking has a long history of strong baselines operating purely on bag-of-words representations ignoring the sequence order entirely. It is therefore all the more interesting to understand the source of the large performance gains \textit{under certain conditions (specifically on MS MARCO)} compared to BM25 (Tab. \ref{tab:baselines}). Is it the potential of the MLMs to model word order?
\label{sec:ranking}
Specifically, we try to answer our first research question:
\begin{description}
\item[\textbf{RQ1}] \sl \rqa
\end{description}

Being able to parse syntax is essential for extracting deeper meaning from natural language. If the model mainly draws on complex language understanding, breaking the sentence structure should lead to a dramatic decrease in performance. 

\subsection{Experiment Design}
We address RQ1 by breaking the order of the input tokens in two different ways: \textit{deterministically sorting} input by token id in decreasing order, and by \textit{random shuffling}. 

The input manipulations can be carried out during fine-tune training and fine-tune evaluation. Forming all combinations of the natural and the perturbed sequence order during training and evaluation results in four coherent experiment conditions for each manipulation.
We are mainly interested in two of those conditions: 
\begin{enumerate}
    \item Perturbing the sequence only zero-shot during evaluation. In this scenario, the model is trained on the original input and perturbed only during evaluation. This tells us how much the fine-tuned ranker is sensitive to changes in the input order. Given the model would be invariant to our input manipulation the performance would not deteriorate compared to the natural order.  
    
    \item Training and evaluating on the same input manipulation. In this setting we allow the model to learn a new representation that is potentially able to cope with unnatural sentence structure in a sense that its representation is entirely invariant to the order.
\end{enumerate}
\begin{table}
    \centering
    \caption{Performance of BM25 (without stemming and default parameters) and the BERT Cross-Encoder (CE) re-ranking the same BM25 ranking on the MSMARCO NIST 2020 testset.}
\begin{tabularx}{0.4\textwidth}{XXXX}
\toprule
model & NDCG@10 & MAP & R@100 \\ 
\midrule
BM25 &  47.96 & 28.56 & 55.99 \\
CE &  68.95 & 45.08 & 68.07 \\
\bottomrule
\end{tabularx}
 \label{tab:baselines}
\end{table}

We manipulate the input by first tokenizing the input. We then apply the respective input manipulation where we only perturb queries and passages within their boundaries.

The position of [SEP] and [CLS] tokens remain unchanged.

\begin{table*}
    \centering
    \caption{Performance of CE with natural-, sorted by descending token-id- and randomly shuffled- input and without position information during different fine-tuning train and fine-tuning eval conditions. All models are trained on MSMARCO, Robust04 serves as a zero-shot evaluation dataset.}
    \label{tab:ce_perturb}
    \small
\begin{tabularx}{1\textwidth}{l|lcc | XXX|XXX}
\toprule 
\multicolumn{4}{c}{}   & \multicolumn{3}{c}{\textbf{MSMARCO}} & \multicolumn{3}{c}{\textbf{Robust04}} \\\cmidrule(lr){5-7}\cmidrule(lr){8-10}
 & model & fine-tune train  & fine-tune eval  & NDCG@10 &  MAP& R@100  & NDCG@10 &  MAP& R@100\\
\midrule

 & CE baseline & natural & natural & 68.95 & 45.08 & 68.07 & 44.27  & 23.31 & 40.35 \\
 
\hdashline

 \multirow{3}{*}{\rotatebox[origin=c]{90}{Sort}} &\multirow{3}{*}{ CE} & natural & sort & 52.36& 32.09 & 64.43  & 39.49 & 21.66& 37.96\\
 &  & sort & sort &67.05 & 43.18 & 69.22  &42.48 & 22.74& 38.45 \\
 & & sort & natural &67.14 &44.05  & 68.85 & 41.23& 22.75&39.25 \\
 
 \midrule
\multirow{3}{*}{\rotatebox[origin=c]{90}{Shuffle}} & \multirow{3}{*}{ CE} & natural & shuffle &  51.78&30.48 & 61.00  &36.66 & 19.41& 35.76 \\
 &  & shuffle & shuffle &  66.71& 42.83& 68.03  & 43.65&22.57 &37.97 \\
 & & shuffle & natural  &65.33 &42.30 & 67.49 & 45.61 & 23.48&38.93 \\
 
 \midrule
 \multirow{3}{*}{\rotatebox[origin=c]{90}{no pos.}} & \multirow{3}{*}{ CE} &  &  &  & &   & & &  \\
 &  & no pos. & no pos. & 66.34  &43.48  & 68.43  &46.22 &24.22  &40.61 \\
 & &  &   & & &  &  & & \\

\bottomrule
\end{tabularx}
\footnotesize
\begin{flushleft}
\end{flushleft}
\end{table*}

\bigskip 
The results of the input manipulation during fine-tune training and fine-tune evaluation can be found in Table  \ref{tab:ce_perturb} for MSMARCO and Robust04. 
We first report the results for the deterministically perturbed input by sorting the input in Sec. \ref{sec:sort} and for shuffling randomly in Sec. \ref{sec:shuffle}. 

\subsection{Results Sorting Deterministically}
\label{sec:sort}
The first line in Tab. \ref{tab:ce_perturb} serves as the baseline where we fine-tune (fine-tune train) and evaluate (fine-tune eval) on the natural sequence order. When we train on the natural sequence order and sort the input we observe a performance drop on all measures for MSMARCO. MAP drops around 30\% (from 45.08 to 32.09), R@100 5\% (from 68.07 to 64.43). For Robust04 we observe a slightly smaller performance drop but follow the same pattern across metrics. The deterioration of the performance on all measures leads to the conclusion that CE is sensitive to the input order by default. Despite the performance drop, the remaining performance shows that the model can cope with destroying the sequence order to some limited extent.

Does this change if we fine-tune the model on the \textit{perturbed input order}?  Surprisingly, when fine-tuning on the sorted input \textit{almost the same performance} as with the natural input (see Tab. \ref{tab:ce_perturb}) can be achieved. The performance is marginally lower for precision-based measures and slightly higher for R@100. This holds for both datasets.
When evaluating this model in turn on the natural sequence order the model's performance drops only marginally showing no clear preference for a particular order, thus being close to entirely order invariant.

\medskip
Regarding RQ1, our findings show that superiority in performance compared to traditional methods can be achieved without drawing upon the natural sequence order. Thus, the ability to model syntactic aspects can be ruled out as a single contributing factor to it. One might argue that the model is performing well on the sorted input because it can reconstruct the original word order, however, \citet{sinha-etal-2021-masked} find evidence to reject this hypothesis.

\subsection{Results Random Shuffling}
\label{sec:shuffle}

Again, the results can be found in Table \ref{tab:ce_perturb}.
The manipulation of the sequence order in the previous experiment is fixed. To avoid picking up artifacts that are based on this particular arrangement such as preserving some syntactical order or contextual information we repeat the experiment with random shuffling. The results can be found as well in Table \ref{tab:ce_perturb}. We find that the perturbation using random shuffling tightly follows the same performance patterns for the different fine-tune and evaluation conditions. The results support our claims made previously that the BERT CE can perform comparably without drawing on the syntactic order of the input. This holds for MSMARCO as well as for Robust04.

\section{Removing Position Information}
\label{sec:nopos}
In this section, we conduct experiments with a new position-ignorant model ``BOW-BERT.''
\medskip

The previous experiments suggest the Cross-Encoder can achieve good performance when fine-tuned on sorted input. While in those experiments the model theoretically still could encode positional information, for this experiment we remove the position information entirely. Specifically, we study our second research question:
\begin{description}
\item[\textbf{RQ2}] \sl \rqd
\end{description}

\subsection{Experiment Design}
For this experiment, we remove the position embeddings that enable the model to perceive input order.  As a consequence, the model is not able to infer the order of the input tokens and can solely revert to a bag-of-words representation of query and document. 

In BERT each input is represented by a combination of three embeddings: token, segment, and position embeddings.
Technically, our new model is a patch of the BERT model that completely removes everything related to the position embeddings. As a result, there is no concept of order over the entire input sequence, and we can no longer rely on order or separator tokens to process the input.  Fortunately, the segment embeddings still allow the model to distinguish which tokens belong to the query or the document (or other token types), so we can still express all crucial information needed to estimate relevance in a ranking setting. 
While this change in the model is technically perhaps small, the conceptual differences are very significant, and we have created a true ``BOW-BERT.''

We first aim to understand how well the model performs without position information on MS MARCO (\ref{sec:ms}) and Robust 04 (\ref{sec:robust}). 
\subsection{Results}
The results can be found at the bottom in Table \ref{tab:ce_perturb}.
\subsubsection{MS MARCO}
\label{sec:ms}
Overall, the performance of the CE without position information is comparable to the CE baseline. The performance drops marginally for precision-based metrics (4\% for NDCG@10, 5\% MAP, and 2\% MRR) compared to the baseline; Recall@100 performance is on par. None of the performance changes is significant with respect to the CE baseline. Our results strengthen the findings that sequence order is not crucial for the performance of the CE to the extent that a bag-of-words representation without any position information can lead to a comparable performance. Interestingly, the model without position information can match the high precision scores of the CE baseline. From a human perspective it is plausible to get a very rough understanding of relevance from a bag-of-words representation, but estimating which of two highly relevant passages is more relevant seems extremely challenging.

\subsubsection{Robust 04}
\label{sec:robust}
The performance of the CE without position information compared to the CE baseline on the Robust 04 dataset can be found in Table \ref{tab:ce_perturb} on the right side. The performance is for all metrics measured comparable. More specific, precision-based metrics are slightly higher without position information: 4\% for NDCG@10, 4\% MAP, and 4\% MRR. 
This small gain may be related to the different requests in Robust (short keyword titles), notably shorter than the fully verbose questions in MS Marco.

\medskip
Regarding RQ2, our findings show that removing position information internally from a BERT cross-coder, a true bag-of-word cross-encoder ranking model, obtains a ranking performance comparable with the standard neural cross-encoder. This finding suggests novel research directions for developing new neural ranking models on more efficient document representations containing e.g. only characteristic words.

\section{Looking into the Models}
\label{sec:analysis}
In this section, we dig deeper and analyse the internal representations of the models, both in general and for the [CLS] token determining the ranking.

\subsection{Inherent Order Sensitivity in Ranking }
The success of CE lies in the pre-training step where representations are learned by masking tokens that the model has to predict correctly. During this task sequence order inarguably plays a role \cite{sinha-etal-2021-masked} and syntax is learned implicitly. Our previous experiments have shown that CE's performance deteriorates but remains relatively high when the input order is manipulated during zero-shot evaluation.
In this section we investigate whether the model becomes more or less order sensitive when we fine-tuned on the ranking task.

\begin{description}
\item[\textbf{RQ3}] \sl \rqb
\end{description}
In other words, how does the order sensitivity of the fine-tuned ranker change compared to the pre-trained BERT model? It might be that the model while being adapted to the ranking task learns to partially ignore word order. We argue losing order sensitivity during fine-tuning would further support the hypothesis that sequence order is not a crucial feature to the CE ranker.

\begin{table}
    \centering
    \caption{CKA similarity between natural and shuffled hidden representations for the vanilla BERT and Cross-Encoder Ranker fine-tuned on natural input order.}
    \label{tab:rq2}
\begin{tabularx}{0.4\textwidth}{XXX}
\toprule
model & fine-tune train & CKA similarity\\
\midrule
BERT & - & 0.4379\\
\hline
\multirow{3}{*}{ CE} & natural & 0.5389\\
 & perturbed & 0.9657\\
\bottomrule
\end{tabularx}
\end{table}

\begin{figure*}
  \centering
  \includegraphics[width=0.45\linewidth]{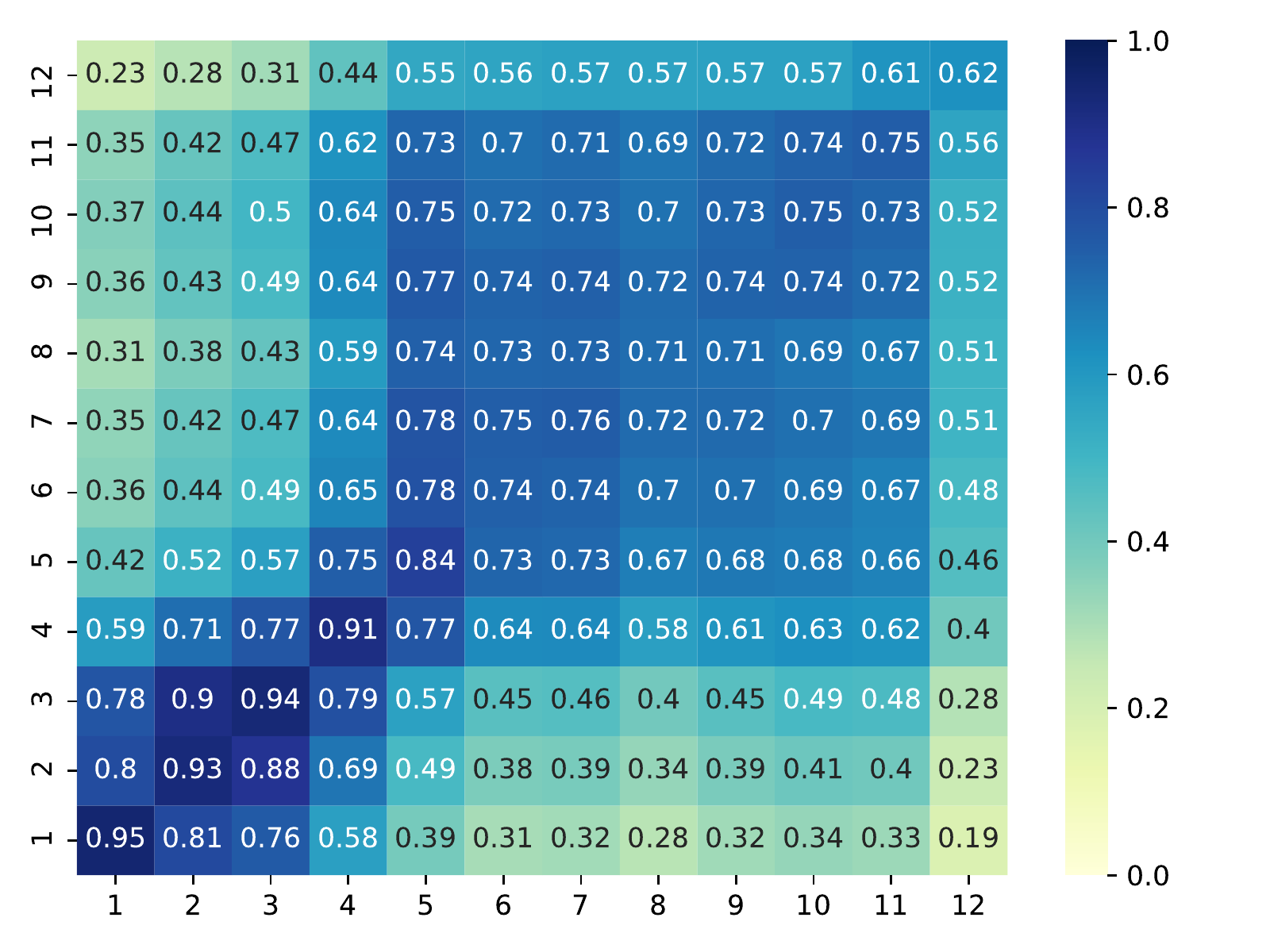}%
    \includegraphics[width=0.45\linewidth]{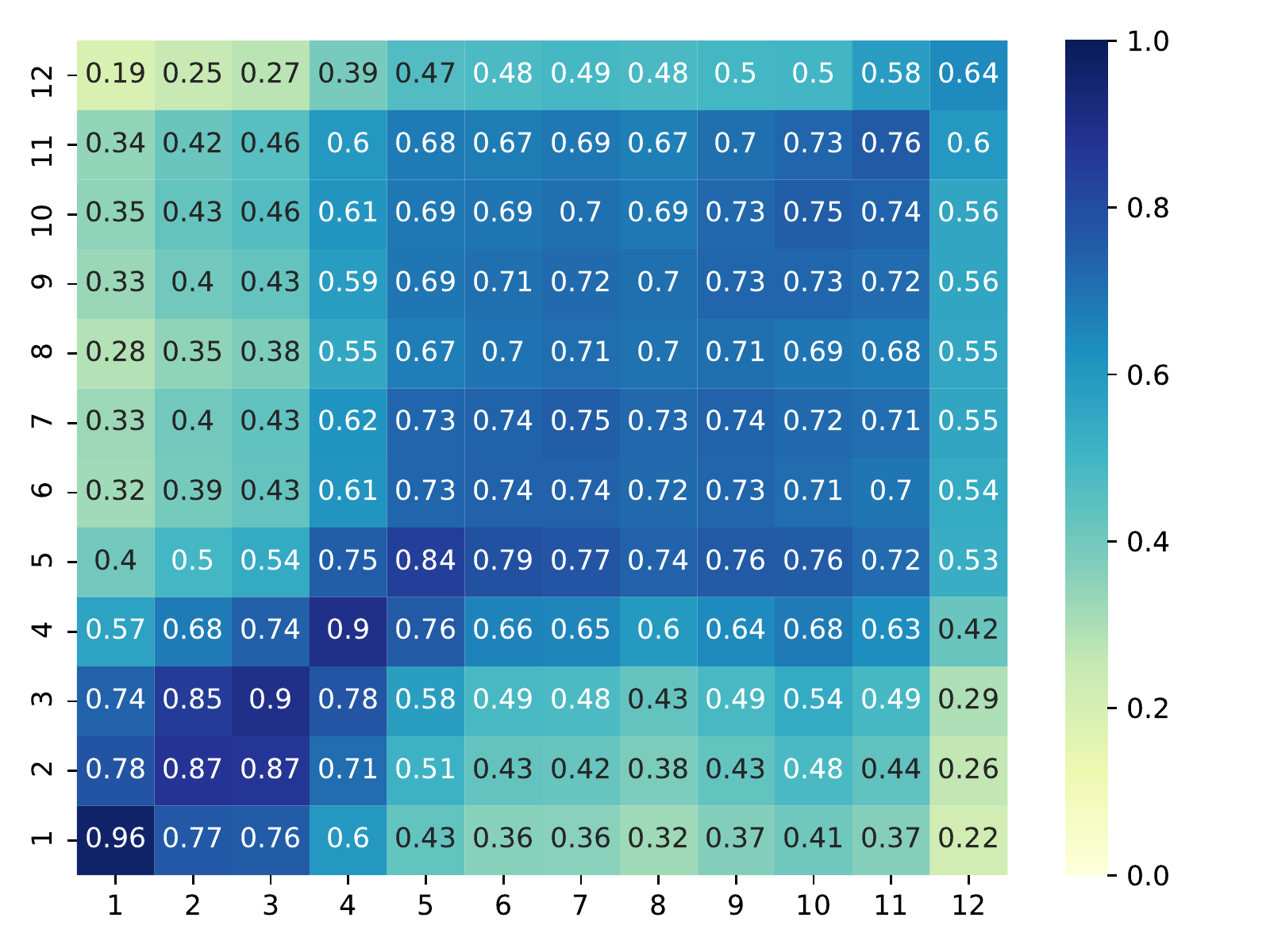}
  \caption{Similarity index kernel-CKA of the hidden representations of different layers measured between the CE baseline and CE trained on sorted input (left) and CE trained without position information (right).}
  \label{fig:cka}
\end{figure*}

\subsubsection{Experiment Design}

We address RQ3 by measuring the difference between the [CLS] representations for the same input with and without perturbing. To quantify differences between the hidden representations we use Centered Kernel-Alignment (CKA) \cite{kornblith} using a linear kernel that can detect meaning similarities between high dimensional representations while being invariant to invertible linear transformations. The similarity is bounded between 0 and 1, where 1 means most similar. We carry out a batch-wise comparison of the representations over the 2020 NIST testset and the average over batches. 

\subsubsection{Results}
We show the CKA similarity between the natural and sorted input for the pre-trained BERT, CE fine-tuned on natural sequence order and CE fine-tuned on sorted order in Table \ref{tab:rq2}. First, we observe that representations of the pre-trained BERT model are most different (CKA similarity 0.4379). Compared to the pre-trained BERT model, the representations of the CE fine-tuned on the natural order are much more similar (CKA similarity 0.5389). We conjecture that the CE fine-tuned ranker fine-tuned on natural input order has lost some of its order sensitivity while being trained on the ranking task compared to its pre-trained base. This confirms our previous presumption in Section \ref{sec:perturbing} that CE partially learns to ignore sequence order during the fine-tuning process. 
The results of CE fine-tuned on the sorted input again confirm that the model is almost entirely invariant (CKA similarity 0.9657) to order perturbations.
\medskip

Regarding RQ3, we conclude that CE is less order sensitive compared to the pre-trained BERT model, suggesting that learning the ranking task leads to focusing less on order sensitivity.

\subsection{Impact on Representations}

Training the model on perturbed input or without position information could lead to a different notion of the input and therefore to different latent representations. We are interested in quantifying how similar the models are compared to the BERT Cross-Encoder ranker. 

\begin{description}
\item[\textbf{RQ4}] \sl \rqc
\end{description}

\subsubsection{Experiment Design}
To quantify the differences between the two models we again apply CKA using a linear kernel. We compare layerwise the kernel-CKA similarity of representations of the CE trained on the natural input to CE trained and evaluated with sorted input. We repeat the same with the model trained without position information. The representations are taken at the end of each layer. We carry out a batch-wise comparison of the representations over the 2020 NIST testset and the average over batches.

\subsubsection{Results}
The results for both models can be found on the left side in Figure \ref{fig:cka}. The comparison of the representations of the CE baseline and CE trained on sorted input shows layers (1-5) exhibit the strongest agreement between the representations of the two models. Representations of layers 6-11 show less agreement while the last layer varies the most. We conclude that training the model on perturbed input order significantly changes the learned representations, especially for later layers. We make the same observations for CE fine-tuned without position information (see Figure \ref{fig:cka} right).
A hypothesis to explain our findings is that word order is being partially maintained within the vanilla CE Ranker as a legacy of the pre-training task causing a drastic change in representations when losing sequence order. 

\medskip
Regarding RQ4, although we observed similar effectiveness, our manipulations destroying sequence order or removing position information result in different latent representations.

\section{Natural Language Understanding}
\label{sec:nlu}
In this section, after having examined the role of modeling syntactic aspects for ranking, we want to revisit NLP tasks under the same conditions. 

\medskip
We are interested in whether our findings are specific for search or whether they are inherent in all BERT-based models.  Hence, our final research question is:  
\begin{description}
	\item[\textbf{RQ5}] \sl \rqe 
\end{description}
To answer this research question we carry out the same input manipulations as done in previous sections for ranking on the GLUE dataset. We fine-tune the model on the natural, sorted after token-id and without position information.

\subsection{Experiment Design}
The GLUE dataset \cite{wang2019glue} is a widely used dataset to evaluate general natural language understanding. It consists of nine different sentence- or sentence-pair language understanding tasks for which a model is fine-tuned separately. We leave out CoLA as this task is to classify whether a sentence is grammatical or not and therefore requires intact sentence structure to be solved. 

We train our BERT models from Hugginface's \cite{ wolf-etal-2020-transformers} \textit{bert-base-uncase} with batch size 64, maximum sequence length 128, and a learning rate 2e-5 for 3 epochs on all tasks, despite for MRPC and WNLI we train for 5 epochs.

\subsection{Results}

The results of comparing fine-tuning on natural input perturbed input, and without position, information can be found in Table \ref{tab:glue}. Note, that the model that is trained on the sorted input is also evaluated on it. Our results show that perturbing the input structure both during fine-tuning and evaluation only yields a marginal decrease from on average 78.2 over all tasks to 78.1 compared to the original input.  Breaking the sentence structure even performs best on tasks RTE, STS-B, and WNLI. 

\begin{table}
    \centering
    \caption{BERT fine-tuned on natural order, sorted by token-id  and without position information on all GLUE tasks, except for CoLA.}
\begin{tabularx}{\linewidth}{lXr@{}lr@{}lr@{}l}  %
\toprule 
Dataset & Metric &  \multicolumn{2}{l}{Natural} &  \multicolumn{2}{l}{Sorted} & \multicolumn{2}{l}{No pos.} \\
\midrule
SST-2 & acc. & 91.6 && 85.6 && 86.0\\
MNLI & acc.  & 84.2 && 79.6 && 79.7\\
MRPC & F1.   & 83.4 && 84.4 && 81.7\\
QNLI & acc.  & 90.7 && 86.6 && 87.1\\
QQP & F1.    & 87.0 &&  85.8 && 85.3\\
RTE & acc.   & 55.2 && 60.2 && 56.6 \\
STS-B & spear. cor & 84.0 && 86.3 && 81.9\\
WNLI & acc.  & 43.6 && 56.3 && 45.0\\
\midrule
All & mean$\pm$std & 78.2 & $\pm$0.19 & 78.1 & $\pm$0.14 & 75.4 & $\pm$0.19 \\
\bottomrule
\end{tabularx}
\footnotesize
\label{tab:glue}
\end{table}

The experiment fine-tuning without position information resembles parts of an experiment conducted by \citet{wang2021on} and \citet{sinha-etal-2021-masked}. We validate their results by finding that removing positional information leads to a small drop (4\%) of the average performance for natural language understanding tasks. 
It is further worth noting that tasks that are based on sentence similarity benefit from the sorted / No position setting: RTE, STS, WNLI.
\medskip

Regarding RQ5, while it is well-known that a BOW representation is effective in IR, our results suggest that syntax also does not seem to play a crucial role to solve the GLUE tasks, consisting of various NLP tasks based on sentence classification.

\section{Conclusion}
In this work, we examined the impact of perturbing the sequence order of the BERT Cross-Encoder re-ranker. We showed that the model, when fine-tuned on destroyed sequence order, can maintain the same performance compared to fine-tuning on natural input. We confirmed this finding by using two different input manipulation that destroys the natural sequence order.
We further find that removing the position information entirely can achieve comparable, if not better, performance to the BERT Cross-Encoder ranker, which is fully informed about the sequence structure of query and document. We confirm our observations based on the evaluation of two different widely used datasets: MS MARCO and Robust 04 (zero-shot document retrieval).

The fact that the models fine-tuned without access to the natural sequence order perform on par with retaining the natural sequence order leads us to the conclusion that syntactic abstraction can not be attributed to performance advantages over earlier models. Our analysis of the hidden representations for the BERT model and the Cross-Encoder ranker further suggests that through fine-tuning the ranking task the sensitivity to the input order is weakened.

Overall, our results point out that syntactic aspects do not play a critical role in the effectiveness of the BERT Cross-Encoder ranker. The performance gain over previous models can not be solely explained by the ability to model sequence order. An explanation for the superior performance of BERT could be the embeddings may be richer representations of meaning regardless of context, and they may represent different word meanings based on aggregated contexts where word order has only a negligible influence. We further conjecture that other aspects such as query-passage cross-attention or deep matching may be contributors to the performance of the model. 

It would be also interesting to observe the effect of perturbing the input and removing the position information during pre-training of BERT. However, due to the computational very demanding pre-training, we have limited our experimentation to fine-tuning only.  Preliminary results suggest that the self-supervised masked word prediction pre-training task is benefiting from word order information, and we may need to rethink the entire training setup. 
We have added to the understanding of the factors responsible for the success of the BERT model mainly by excluding plausible explanations mirroring human text understanding.  This immediately suggests novel research directions to directly address the factors that are key in the BERT model.  Our findings give rise to the presumption that the current pre-training task of the underlying BERT model is over-complex for ranking and other downstream classification tasks.  Implicitly promoting to model word order and syntactic aspects could potentially take up a large capacity of the model. Our findings provide a good starting point for the design of pre-training tasks tailored specifically to ranking, possibly reducing complexity and model size.

As a general observation, the effectiveness of recent transformer-based rankers has frequently been characterized as ``bringing NLP into IR'' because they present a departure from the text statistics-based bag-of-word models.  Our analysis may suggest that in a way these models are ``bringing IR into NLP'' showcasing that (higher-order) word statistics even solves NLP problems traditionally thought to require the representation of complex syntactic structures.

\begin{acks}
This research is funded in part by 
the Netherlands Organization for Scientific Research %
(NWO CI \# CISC.CC.016), and 
the Innovation Exchange Amsterdam (POC grant).
Views expressed in this paper are not necessarily shared or endorsed by those funding the research.
\end{acks}


\end{document}